\documentclass[a4paper,fleqn]{cas-dc}

\usepackage[authoryear,longnamesfirst]{natbib}
\usepackage{xcolor}
\usepackage{float}
\def\tsc#1{\csdef{#1}{\textsc{\lowercase{#1}}\xspace}}
\tsc{WGM}
\tsc{QE}
\tsc{EP}
\tsc{PMS}
\tsc{BEC}
\tsc{DE}

\begin{document}
\let\WriteBookmarks\relax
\def\floatpagepagefraction{1}
\def\textpagefraction{.001}

\shorttitle{Phrase Mining}

\title [mode = title]{Language Model as an Annotator: Unsupervised Context-aware Quality Phrase Generation}     

\author[1]{Zhihao Zhang}[
                        ]
                        
\author[2]{Yuan Zuo}[
                        orcid=0000-0001-8516-2567
                        ] \corref{cor1} 
\cortext[cor1]{Corresponding author}
\ead{zuoyuan@buaa.edu.cn}

\author[3]{Chenghua Lin} [
                        ] 
                        
\author[2]{Junjie Wu} [
                        ] 

\affiliation[1]{organization={College of Economics and Management, Beijing University of Technology},
    city={Beijing},
    country={China}}
    
\affiliation[2]{organization={School of Economics and Management, Beihang University},
    city={Beijing},
    country={China}}

\affiliation[3]{organization={Department of Computer Science, University of Manchester},
    city={Manchester},
    country={UK}}

\begin{abstract}
Phrase mining is a fundamental text mining task that aims to identify quality phrases from context. Nevertheless, the scarcity of extensive gold labels datasets, demanding substantial annotation efforts from experts, renders this task exceptionally challenging. Furthermore, the emerging, infrequent, and domain-specific nature of quality phrases present further challenges in dealing with this task. Therefore, in this paper, we propose LMPhrase, a novel unsupervised context-aware quality phrase mining framework built upon large pre-trained language models (LMs). Specifically, we first mine quality phrases as silver labels by employing a parameter-free probing technique called Perturbed Masking on the pre-trained language model BERT (coined as \textit{Annotator}). In contrast to typical statistic-based or distantly-supervised methods, our silver labels, derived from large pre-trained language models, take into account rich contextual information contained in the LMs. As a result, they bring distinct advantages in preserving informativeness, concordance, and completeness of quality phrases. Secondly, training a discriminative span prediction model heavily relies on massive annotated data and is likely to face the risk of overfitting silver labels. Alternatively, motivated by recent success in formulating language understanding problems such as named entity recognition and sentiment analysis as generation tasks, we formalize phrase tagging task as the sequence generation problem by directly fine-tuning on the Sequence-to-Sequence (Seq2Seq) pre-trained language model BART with silver labels (coined as \textit{Generator}). Finally, we merge the quality phrases from both the Annotator and Generator as the final predictions, considering their complementary nature and distinct characteristics. Extensive experiments show that our LMPhrase consistently outperforms all the existing competitors across two different granularity phrase mining tasks, where each task is tested on two different domain datasets. The promising results show the superiority of our framework with pre-trained language model.
\end{abstract}


\maketitle

\section{Introduction} \label{introduction}

Quality phrase refers to informative multi-word expression that ``\emph{appears consecutively in the text, forming a complete semantic unit in certain contexts or the given document}''~\citep{finch2016linguistic}. Phrase mining task aims to recognize quality phrases from context. As an effective technique to transform unstructured text into structured information, phrase mining has immense value to various downstream natural language processing tasks including named entity recognition~\citep{DBLP:conf/emnlp/ShangLGRR018,peng2021named}, topic modeling~\citep{DBLP:journals/pvldb/El-KishkySWVH14}, and taxonomy construction~\citep{DBLP:conf/kdd/ShenWLZRVS018}, to name a few. In short, phrase mining plays a fundamental role in facilitating human processing and understanding of the unstructured text data.

In the field of phrase mining, statistic-based approaches have long been widely studied as the mainstream methods. Two representative approaches are  SegPhrase~\citep{DBLP:journals/pvldb/El-KishkySWVH14} and AutoPhrase~\citep{DBLP:journals/tkde/ShangLJRVH18}. Typically, these approaches frame the task of phrase mining as a binary classification problem. They leverage various frequency signals as features to classify candidate phrase into either high-quality or non-quality phrase. Such approaches that rely on statistical signals have yielded very good results. However, recent research~\citep{DBLP:conf/sdm/AnjumAXH21} points out that both quality and noisy phrases follow the power-law distribution. The noisy phrases consistently appear alongside quality phrases throughout frequency spectrum, which makes it difficult to separate quality phrases from noisy ones using statistic-based methods. To tackle this issue, DsPhrase~\citep{DBLP:conf/cikm/WangZJZWNXX20} formulates phrase mining task as a sequence labeling problem and then incorporates implicit semantic features to estimate the quality of the phrase. However, DsPhrase uses knowledge bases (KB) to string-match a corpus for obtaining distant supervision signals~\citep{DBLP:conf/emnlp/ShangLGRR018}, where the partially matched phrase might introduce bias into the phrase mining model.

To alleviate the manual annotation efforts by domain and linguistic experts and eliminate the bias introduced by distantly-supervised signals, UCPhrase~\citep{DBLP:conf/kdd/GuWBML0S21} explores phrase tagging in an unsupervised manner. Specifically, quality phrases are induced as silver labels by extracting maximal word sequence from the given document. Then, a transformer-based span classification model is proposed to identify quality phrases from the sentence. Despite their success, it's important to note that the silver label acquisition algorithm remains fundamentally rooted in statistical methods, which could potentially lead to the omission of infrequent quality phrases. In addition, training a neural discriminative span classifier heavily depends on an extensive volume of labeled data, which can make it susceptible to overfitting with silver labels. In conclusion, it is challenging to derive high quality silver labels in a context-aware manner and to train a robust prediction model that can withstand noises. Thus, how to recognize quality phrases accurately and efficiently in an unsupervised manner still remains an challenge.

To tackle the aforementioned challenges, we propose LMPhrase, a novel unsupervised context-aware quality phrase mining framework with large pre-trained language model. Our LMPhrase is a two-stage framework: we first resort to auto-encoding pre-trained language model BERT as \textbf{Annotator} to mine high-quality silver labels from the sentence by employing a perturbed masking technique. Given the silver labels, we then build \textbf{Generator} by directly fine-tuning on the encoder-decoder pre-trained language model BART as a sequence generation task to generate quality phrases. Lastly, we merge the quality phrases from both Annotator and Generator as the final predictions, considering the complementary nature of Annotator and Generator due to their distinct characteristics. Our work is not just a combination of BERT (\textbf{Annotator}) and BART \textbf{Generator} models. Our framework is built upon large pre-trained language models (LMs), which is completely different from the previous ones. The reasons are as follows:

The large language models pre-trained on massive text corpora, such as BERT~\citep{DBLP:conf/naacl/DevlinCLT19}, have been shown to encode a great deal of knowledge in their parameters implicitly. Inspired by the parameter-free probing technique known as Perturbed Masking, which is used for the analysis and interpretation of pre-trained language models~\citep{DBLP:conf/acl/WuCKL20}, we introduce this probing technique into masked language modeling objective to measure the effect of a word $x_j$ on predicting another word $x_i$ in this paper. This inter-word information, denoted as impact matrix $\mathcal{F}$, contains inter-word correlations and forms the foundation for generating silver labels. We then design a heuristic segmentation algorithm to derive silver labels from this impact matrix directly. Note that it is quite different from utilizing BERT as a contextual feature extractor, our Annotator is derived from analyzing and interpreting BERT, which enjoys huge syntactic, semantic and structural knowledge from large pre-trained language model. Therefore, compared to context-agnostic statistic-based or distantly-supervised approaches, our silver labels take into account the contextual information, and thus offer the advantages in preserving informativeness, concordance, and completeness of quality phrases.

Given the high quality silver labels derived from our automatic Annotator, we design a tailored model to further recognize quality phrases. Different from a number of phrase mining approaches that involve intricate model architectures, specialized statistical signals, and semantic features for training discriminative span prediction models, we show that a simple Seq2Seq pre-trained language model, BART~\citep{DBLP:conf/acl/LewisLGGMLSZ20}, can be easily adapted as Generator to generate quality phrases. Specifically, we formalize the phrase tagging task as a Seq2Seq sequence generation problem by directly fine-tuning on the pre-trained model BART using its conditional generation method typically used for text summarization. To this end, we define the source ($S$) and target ($T$) sequences, respectively. Note that $S$ represents source input sentence. We organize $T$ by concatenating all quality phrases mined by Annotator into a single string according to their occurrence order in the source sentence, where comma is used as special token for splicing quality phrases. In this way, the heavy dependence of annotations and the risk of overfitting are mitigated. We also demonstrate that, even with simple and standard fine-tune procedure with limited annotated data (silver labels), our Generator can yield competitive results on benchmarks compared to existing strong baselines.

It is worth mentioning that even human annotators may not always reach a unanimous consensus on some specific phrases, which can result in a situation where that diverse phrases, including those with some overlap, are all marked as golden quality phrases. In addition, our Annotator and Generator have distinct characteristics. In particular, Annotator is a purely unsupervised model derived from pre-trained language model BERT, which can mine common patterns or collocations contained in the large language model. In contrast, Generator is a fully supervised model trained with the silver labels, where the pre-trained language model BART plays a key role. All of these motivate us to merge the quality phrases induced from both Annotator and Generator as the final prediction. The experimental observation also shows the complementary nature of Annotator and Generator, meaning that Annotator achieves high accuracy but relatively low recall, while the opposite is true for Generator. Thus, the merged results contain more diverse and accurate quality phrases than the individual component.

In summary, we propose a innovative unsupervised context-aware quality phrase mining framework that leverages large pre-trained language model to overcome the challenge posed by the absence of large-scale gold labels in text mining tasks. Our framework benefits from extensive syntactic, semantic, and structural knowledge from large pre-trained language models, and thus has a clear advantage in mining low-frequency, emergent, and domain-specific phrases. Experiments on sentence-level phrase tagging and document-level keyphrase extraction tasks on different domain datasets also show the superiority of our LMPhrase over all existing strong competitors. 

The main contributions of our work are summarized below:
\begin{enumerate}
	\item To the best of our knowledge, LMPhrase is the first attempt to analyze, interpret and utilize the pre-trained language model BERT for unsupervised context-aware quality phrase mining by employing Perturbed Masking technique on it.
    \item We formalize phrase tagging task as a Seq2Seq sequence generation problem by fine-tuning on pre-trained language model BART to generate quality phrases, where the heavy dependence of annotations and the risk of overfitting are mitigated.
    \item Our approach consistently outperform all the state-of-the-art competitors across two different granularity phrase mining tasks, each with two domains datasets. We also analyze and discuss each model component in detail.
\end{enumerate}

The rest of our paper is organized as follows.  Section~\ref{related_work} reviews the related work of this paper. Section~\ref{preliminaries} briefly describes the preliminaries as the background knowledge Section~\ref{methodology} illustrates a detailed explanation of our LMPhrase framework. Section~\ref{exp_setup} describes our experimental setups, which is followed by the experimental results in Section~\ref{exp_result}. We finally conclude the paper in Section~\ref{conclusion}.
\section{Related Work} \label{related_work}

\subsection{Phrase mining}
Phrase mining refers to recognize semantically meaningful multi-word span as a whole semantic unit from the given corpora. There has been extensive research in studying and designing effective methods to accurately mine quality phrases from the unstructured text. Early works primarily center on conventional statistic-based approaches, which heavily rely on statistical signals derived from large-scale corpora to identify quality phrases~\citep{DBLP:conf/acl/Deane05, DBLP:conf/dasfaa/LiW0YL16, DBLP:conf/aaai/LiYWC17}. Most of them leverage various statistical features derived from document collections to estimate the quality of candidate phrase. \cite{DBLP:journals/pvldb/ParameswaranGR10} propose several indicators, including frequency and comparison with super/subsequences, to extract quality phrases. \cite{DBLP:journals/pvldb/El-KishkySWVH14} employ t-statistic to filter and rank the candidate phrases. \cite{DBLP:conf/sigmod/LiuSWRH15} enhance the estimation of phrase quality with phrasal segmentation to further correct the initial set of the statistical features. Unlike the purely unsupervised methods described above, SegPhrase~\citep{DBLP:journals/pvldb/El-KishkySWVH14} depends on the gold labels annotated by domain experts to train the phrase quality estimator. AutoPhrase~\citep{DBLP:journals/tkde/ShangLJRVH18} resorts to knowledge bases (KB) based distantly-supervised approach to mine quality phrases, which can be used in many different domains with a minimum of human effort.

All of the above models either are tightly coupled to frequency statistics to measure the quality of phrases, which is insufficient to distinguish quality phrases from the noises, or rely heavily on domain experts for large-scale manually annotated gold labels. To address this issue, UCPhrase~\citep{DBLP:conf/kdd/GuWBML0S21} proposes an unsupervised phrase tagging framework, where quality phrases are induced as silver labels by extracting maximal word sequences from the given document, and then a transformer-based span classification model is trained to predict quality phrases. Although this work yields encouraging performance on the benchmark dataset, silver label generation algorithm is still essentially statistic-based, and training a neural discriminative span classifier heavily relies on massive annotated data and faces the risk of overfitting. Alternatively, we propose LMPhrase, a novel unsupervised context-aware quality phrase mining framework with pre-trained language model. Our silver labels Annotator considers the contextual information, and thus offer the advantages in preserving informativeness, concordance, and completeness of quality phrases. Our quality phrase Generator is more robust to the noise and able to generate quality phrases without specifically designed architectures.

\subsection{Investigation of Pre-trained Language Model}
Recent prevalent large pre-trained language models (LMs), such as ELMO~\citep{DBLP:conf/naacl/PetersNIGCLZ18}, BERT~\citep{DBLP:conf/naacl/DevlinCLT19}, and GPT-3 ~\citep{DBLP:conf/nips/BrownMRSKDNSSAA20}, have been shown to encode a great deal of knowledge within their parameters implicitly. This has sparked extensive research delving into investigating what the pre-trained language models know. One emerging research direction is the use the of probing task to investigate different syntactic properties and linguistic knowledge captured by the model~\citep{DBLP:conf/blackboxnlp/ClarkKLM19, DBLP:conf/iclr/TenneyXCWPMKDBD19, DBLP:conf/acl/JawaharSS19}. Among all of these related probe tasks, \cite{DBLP:conf/acl/WuCKL20} propose a parameter-free probing technique called Perturbed Masking to analyze and interpret pre-trained language models. Its core idea is to introduce this probing technique into the masked language modeling objective to measure inter-word correlations. This inter-word correlation value indicates the impact of a word $x_j$ on predicting another word $x_i$. Motivated by this, we hypothesize that the higher correlation value, the stronger the association between these two adjacent words, and the more likely they are to form a complete phrase. To this end, we design a heuristic segmentation algorithm to derive quality phrases as silver labels from inter-word correlations. Note that our silver labels mined from pre-trained language models enjoy rich knowledge and take contextual information into account, and thus offer the advantages in preserving informativeness, concordance, and completeness of quality phrases.

\subsection{Sequence-to-Sequence Models}
The Seq2Seq learning framework has been studied for a long time and is widely adopted in Natural Language Processing (NLP) Community~\citep{DBLP:conf/nips/SutskeverVL14, DBLP:conf/emnlp/LuongPM15, DBLP:conf/nips/VaswaniSPUJGKP17,li-etal-2020-dgst,DBLP:journals/kbs/XiaoLYGXL21}. Recently, the impressive performance gains achieved by pre-trained language models has prompted attempts to pre-train the Seq2Seq model~\citep{DBLP:conf/acl/LewisLGGMLSZ20, DBLP:journals/jmlr/RaffelSRLNMZLL20}. Among them, BART~\citep{DBLP:conf/acl/LewisLGGMLSZ20}, a strong Seq2Seq generative pre-trained model, is employed for a multitude of downstream Natural Language Generation (NLG) tasks, including text summarization~\citep{DBLP:journals/corr/abs-2010-09252,goldsack-etal-2022-making,tu2022pcae}, name entity recognition~\citep{DBLP:conf/acl/YanGDGZQ20}, aspect-based sentiment analysis~\citep{DBLP:conf/acl/YanGDGZQ20, zhang2022aspect}, and keyphrase generation~\citep{DBLP:journals/corr/abs-2201-05302, DBLP:journals/kbs/YangGYY22}. Inspired by the literature above, we investigate whether the standard Seg2Seq generative architecture can be used for sentence-level phrase tagging. To this end, we formalize the phrase tagging task as the Seq2Seq sequence generation problem by directly fine-tuning on the pre-trained language model BART. It is noteworthy that our generative model achieves comparable results and is robust to the silver labels derived from the Annotator. The experimental results also demonstrate that, even with simple and standard fine-tune procedure, our generative model with limited annotated data can achieve results on par (in some cases or even better) with the strongest competitors that have specifically designed architectures.
\section{Preliminaries} \label{preliminaries}

Before delving into the specific design of our framework, we first give a clear definition of quality phrase and present the statistics of two manual annotated sentence-level phrase tagging datasets as background knowledge. Then we formulate the research problem and briefly describe the problem definition. We also provide all the notations, terminologies and abbreviations used in this paper for reference purpose in Table~\ref{tab:notations}.

\subsection{Quality Phrase} \label{quality_analysis}
Quality phrase refers to informative multi-word expression that ``\emph{appears consecutively in the text, forming a complete semantic unit in certain contexts or the given document}''~\citep{finch2016linguistic}. Generally, a quality phrase should meet the criteria as below~\citep{DBLP:conf/sigmod/LiuSWRH15}:
\begin{itemize}
	\item \emph{Popularity}: Quality phrase should appear with sufficient frequency in a given document collection.
    \item \emph{Concordance}: Concordance refers to the collocation of tokens in such frequency that is significantly higher than what is expected due to chance.
	\item \emph{Informativeness}: A phrase is informative if it is indicative of a specific topic.
	\item \emph{Completeness}: A phrase is deemed complete when it can be interpreted as a whole semantic unit in certain context.
\end{itemize}

\begin{table}[htbp]
	\centering
    \caption{Statistics of two manual annotated sentence-level phrase tagging datasets. \#Sents denote the numbers of sentences. \#AveWords and \#Avephrase represent the averaged numbers of words and quality phrases per sentence.}
	\resizebox{0.9\linewidth}{!}{
		\begin{tabular}{c|c|c|c|c} \hline
			Datasets & domain & \#Sents & \#AveWords & \#Avephrase  \\ \hline
			KP20k  & Scientific & 200 & 25 & 2.3    \\ 
			KPTimes & News & 200 & 24 & 1.5 \\ \hline
	\end{tabular}}
	\label{sent_statics}
\end{table}

We analyze two manually annotated sentence-level phrase tagging datasets to further interpret the definition of quality phrase~\citep{DBLP:conf/kdd/GuWBML0S21}. Table~\ref{sent_statics} shows the statistics of these two manual annotated datasets. Note that the annotator was asked to tag the multi-word spans which are very likely to be the keyword for the given sentence. The quality phrase is annotated by three annotators independently, and the agreement between human annotations is $90\%$.

We also analyze the part-of-speech (POS) tag patterns of these manual annotated quality phrases. Concretely, we first use StanfordCoreNLP Tools\footnote{https://stanfordnlp.github.io/CoreNLP/} to obtain POS tags of the input sentence, then we extract and aggregate the patterns of all annotated quality phrases. It can be observed that most of the quality phrases are noun phrases\footnote{A phrase which consists of zero or more adjectives followed by one or multiple nouns is called Noun Phrase.}. In addition, as shown in Table~\ref{tab:phrase}, another empirical observation is that the quality phrases could be mainly classified into four types, which also meets the above criteria. These observations suggest that this manual annotated dataset can serves as a good test bed for sentence-level phrase tagging task, and on the other hand motivate our specific design of the model for this task. We will cover these in detail in the next sections.

\begin{table}[]
	\centering
    \caption{Classification of quality phrases.}
    \begin{tabular}{cc}
    \hline
        Types & Examples \\ \hline
        Named Entity & Nissan Motor Co. \\ 
        Terminology/Concept & Homology Theory \\ 
        Emerging Phrase & ME Size \\ 
        Acronyms & TFP Growth \\ 
        ... & ... \\ \hline
\end{tabular}
\label{tab:phrase}
\end{table}

\subsection{Problem Definition}
Given a sentence with $t$ words $[ x_1, x_2, .., x_t]$, the quality phrase is a sequence of consecutive words $[x_i, ..., x_{i+k}]$ that form a complete and informative semantic unit in context. We follow the setting of UCPhrase~\citep{DBLP:conf/kdd/GuWBML0S21}, considering multi-word phrases, which tend to carry more information, as potential quality phrases. This setup is considered more challenging due to both diversity and sparsity.

In this paper, we mainly focus on two different granularity phrase mining tasks: sentence-level phrase tagging and document-level keyphrase extraction, covering two different domains datasets. Sentence-level phrase tagging is a fine-grained task designed to extract quality phrases for each input sentence. Document-level keyphrase extraction is a relatively coarse-grained task that aims to extract keyphrases that can best summarize the given article. More details please refer to Section~\ref{exp_setup}.

\begin{table}[]
\caption{Notations, terminologies and abbreviations}
\label{tab:notations}
\centering
\resizebox{0.95\linewidth}{!}{
\begin{tabular}{cc}
\hline
Notation/Terminology/Abbreviations & Description \\ \hline
Knowledge Base &  a centralized repository of information  \\  
Sequence-to-Sequence & a family of machine learning approaches \\
TF-IDF & term frequency - inverse document
frequency \\
POS & part-of-speech \\
NLP & natural language processing \\
NLG & natural language generation \\
LMs & pre-trained language models \\ \hline
\textbf{s}  &  input sentence  \\ 
\textbf{t}  &  quality phrase sequence  \\ 
$X$ & source sequence \\
$Y$ & target sequence \\
$x_i$  &  input token i  \\ 
$[M]$ & [MASK] token \\
$H_\theta(x)_i$ & contextual representation of token $x_i$ \\ 
$\textbf{s}\backslash\{x_i\}$ & sequence with $x_i$ replaced by [MASK] \\
$H_\theta(\textbf{s}\backslash\{x_i\})_i$ & representation of $x_i$ replaced by [MASK] \\
$f(x_i, x_j)$ & distance function \\
$d(x, y)$ &  euclidean distance \\
 $\mathcal{F}$ & impact matrix \\
 $\theta$ & the network parameters of BERT \\
 $P(Y|X)$ & loss function of Generator\\
 $F1@10$ &  $F1$ score of the top-10 ranked phrases \\
\hline
\end{tabular}}
\end{table}
\section{Methodology} \label{methodology}

\begin{figure*}[t!]
    \centering
    \includegraphics[scale=0.4]{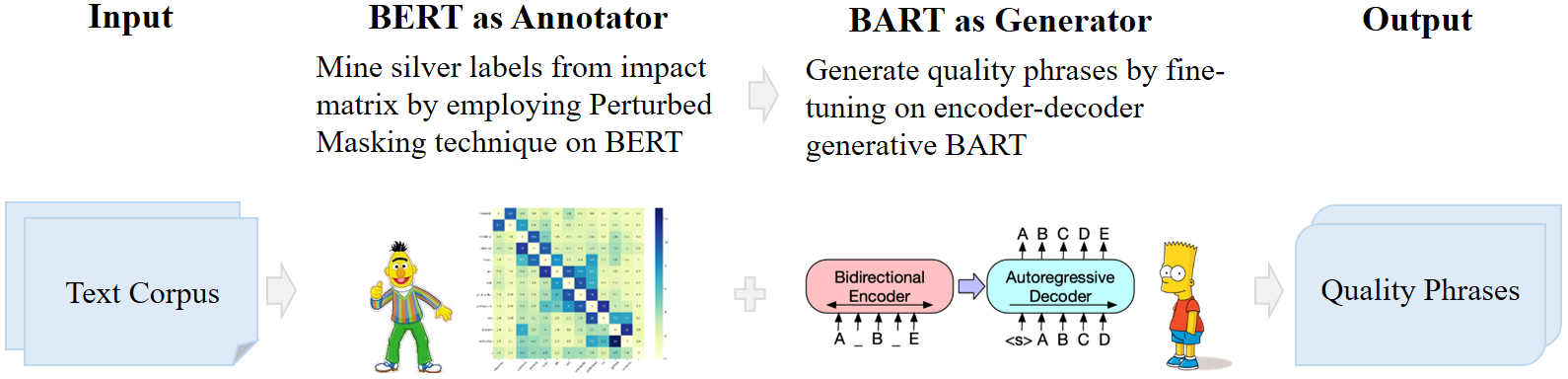}
    \caption{The architecture of LMPhrase: An unsupervised context-aware quality phrase mining framework with pre-trained language model.}
    \label{fig:framework}
\end{figure*}

In this section, we describe our LMPhrase framework in detail for unsupervised context-aware quality phrase mining with pre-trained language model. LMPhrase first adopts pre-trained language model BERT as Annotator to mine quality phrases from each sentence as silver labels. Given the silver labels, the pre-trained language model BART is then employed as Generator to recognize the quality phrases directly. Lastly, we merge the quality phrases from both Annotator and Generator as the final predictions. An overview of our LMPhrase is shown in Figure~\ref{fig:framework}.

\subsection{Annotator: Silver Label Acquisition}
In the first step, we resort to pre-trained language model BERT as Annotator to mine quality phrases from context in an unsupervised manner, and these quality phrases will be served as silver labels for the training of our following Generator. To tackle the problem, we introduce Perturbed Masking technique~\citep{DBLP:conf/acl/WuCKL20} into the masked language modeling (MLM) objective to estimate inter-word correlations, which are called Impact Matrix. We then design a heuristic segmentation algorithm to induce the quality phrases as silver labels from this inter-word correlation information.

\subsubsection{Token Perturbation}
Given an input sentence $\textbf{s}=[x_1, x_2, ..., x_t]$ of length $t$, we employ BERT to map each token $x_i$ into the contextual representation $H_\theta(x)_i$, where $\theta$ denotes the network parameters of BERT. We adopt token Perturbed Masking technique to capture the effect of a word $x_j$ on predicting another word $x_i$ within the sentence. 

\begin{figure}[htbp]
    \centering
    \includegraphics[scale=0.4]{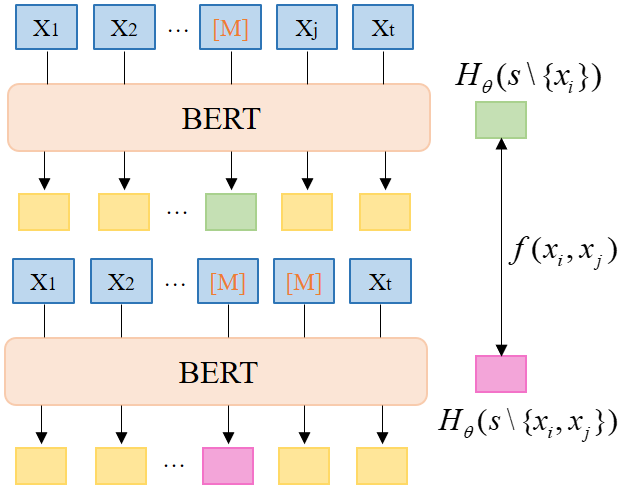}
    \caption{Illustration of token perturbation.}
    \label{fig:perturb}
\end{figure}

Figure~\ref{fig:perturb} illustrates the token perturbation process. Specifically, we first replace word $x_i$ with [MASK] token and then feed this masked sequence $\textbf{s}\backslash\{x_i\}$ into BERT. $H_\theta(\textbf{s}\backslash\{x_i\})_i$ denotes the representation of
$x_i$ (i.e. the representation of the [MASK] token). To calculate the impact that $x_j$ has on $x_i$, we further mask $x_j$ to obtain another masked sequence $\textbf{s}\backslash\{x_i, x_j\}$. Similarly, we use $H_\theta(\textbf{s}\backslash\{x_i, x_j\})_i$ to denote the new representation of token $x_i$. We define a function $f(x_i, x_j)$ to measure the effect of a word $x_j$ on predicting another word $x_i$ as below:
\begin{equation}
    f(x_i, x_j) = d(H_\theta(\textbf{s}\backslash\{x_i\})_i, H_\theta(\textbf{s}\backslash\{x_i, x_j\})_i), 
\end{equation}
where $d(\textbf{x}, \textbf{y})$ is the Euclidean Distance between the vector representations of $x_i$ and $x_j$, which essentially measures the correlation between them. We also experiment with other metrics such as cosine similarity and KL divergence. But we don't not observe significant differences.

\subsubsection{Impact Matrix}
By repeating this two-stage token perturbation masking calculation process on each pair of words $x_i, x_j \in \textbf{s}$, we can obtain an impact matrix $\mathcal{F}$, where $\mathcal{F}_{ij} \in \mathbb{R}^{t \times t}$. It is noteworthy that that BERT adopts byte-pair encoding~\citep{DBLP:conf/acl/SennrichHB16a} and could split a word into multiple tokens (subwords). To derive the impact matrix on the word-level, we mask all the tokens of a split-up word in each perturbation. The impact on a split-up word is obtained by averaging\footnote{We also experimented with other alternatives (e.g. max poling), but observe no significant difference.} the impacts over the split-up word’s tokens.

\begin{figure}[h]
    \centering
    \includegraphics[scale=0.22]{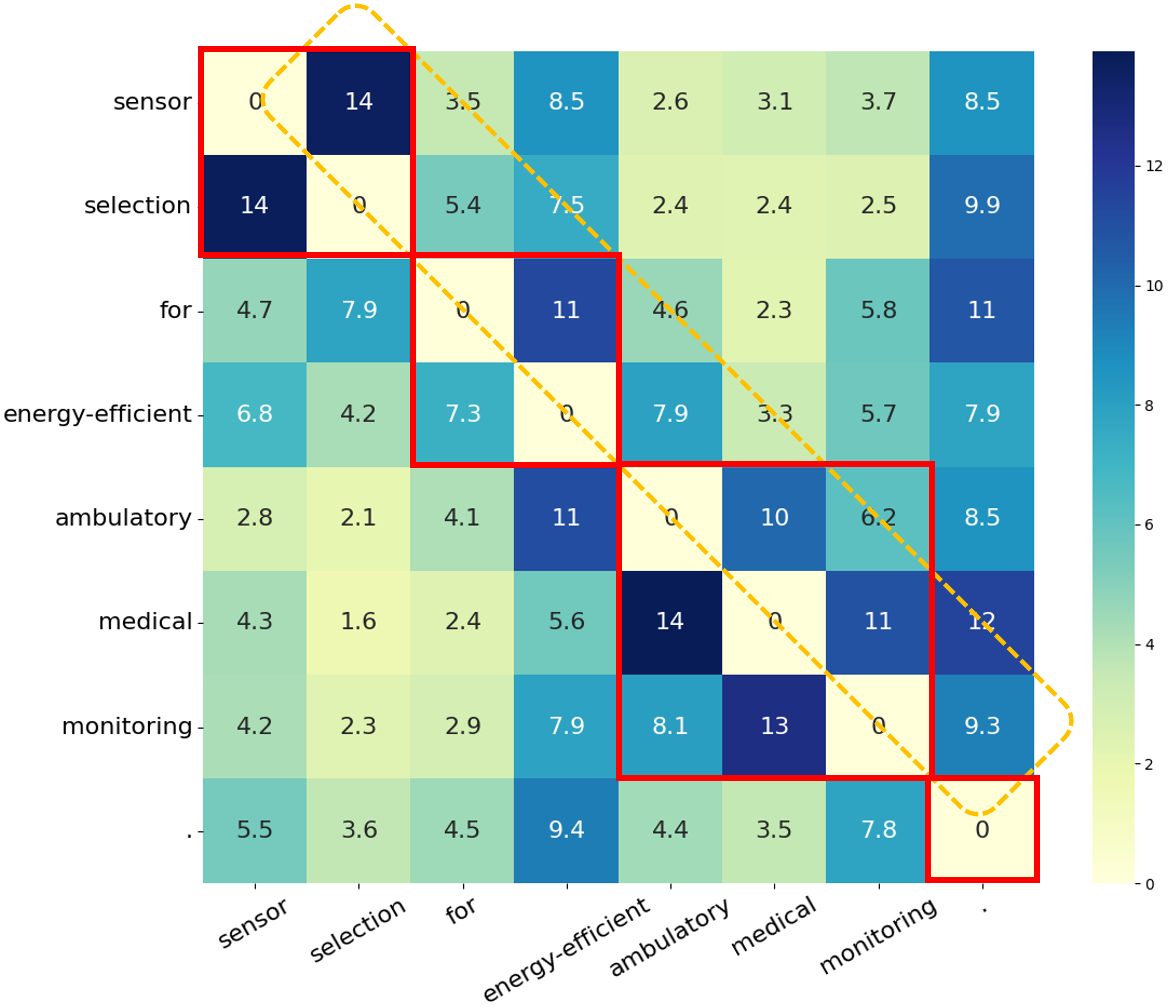}
    \caption{Visualization of the Heatmap for the sentence ``sensor selection for energy-efficient ambulatory medical monitoring.”}
    \label{fig:impact_matrix}
\end{figure}

Before proceeding to silver label mining, we first visualize the impact matrix of a sample sentence with a heatmap. The heatmap of the impact matrix for the  sentence ``\textit{sensor selection for energy-efficient ambulatory medical monitoring .}" is shown in Figure~\ref{fig:impact_matrix}. The values of impact matrix reflect the inter-word correlations between any two words. Take the word “monitoring” as an example, we can see a clear vertical stripe above the main diagonal, which means that this particular occurrence of the word ``monitoring'' strongly influences the occurrences of the words proceeding it. In other words, ``monitring'' and  its proceeding tokens are likely to form a quality phrase due to their tie correlation. Such observations motivate us to explore the idea of mining phrase from this impact matrix in an unsupervised manner. The impact matrix derived from the given sentence serve as the basis for our later heuristic segmentation algorithm.

\subsubsection{Segmentation Algorithm} \label{segmentation}
Given the impact matrix derived from token perturbation process, we design a heuristic segmentation algorithm to mine quality phrases. Specifically, we first select the top right diagonal values (i.e. above the main diagonal) as shown in the yellow dashed box of Figure~\ref{fig:impact_matrix}, which represent the correlations from all pairs of two adjacent words. We hypothesize that the higher value indicates the stronger association between two adjacent words, which are more likely to form a complete phrase. We determine whether two adjacent words can form a phrase by setting a threshold, that is, if the correlation value of two adjacent words is greater than the threshold, it shows that the two adjacent words are closely related and are likely to form a phrase; otherwise, the two adjacent words are less likely to form a phrase and should be separated. In this way, the sentence is split into four chunks as shown in the red box of the Figure~\ref{fig:impact_matrix}, which serve as four potential candidate phrases.

To exclude uninformative candidate phrases (e.g., “for energy-efficient”), We first filter out the stopwords before determining the final silver labels. Uninformative words are excluded with the stopword list commonly used in previous work~\citep{DBLP:conf/sigmod/LiuSWRH15, DBLP:journals/tkde/ShangLJRVH18}. Based on our observation, the majority of quality phrases are noun phrases, as discussed in Section~\ref{quality_analysis}. Therefore, we further conduct regular expression matching to filter candidate phrases which are not noun phrases. We treat the remaining candidate phrases as quality phrases as silver labels for later Generator training. Note that although our language model based Annotator is purely an unsupervised model that does not require any manually annotated data, it demonstrates very good results especially for the precision measure on sentence-level phrase tagging task (More details please refer to Section~\ref{Silver_Labels}).

\subsection{Generator: Quality Phrase Generation}
In general, training a discriminative span prediction model heavily relies on massive annotated data and usually faces the risk of overfitting with silver labels. Alternatively, inspired by the prior work~\citep{DBLP:conf/naacl/PetroniPFLYCTJK21} that a multitude of downstream NLP tasks can be effectively addressed using a Seq2Seq generative framework, we examines whether a standard Seq2Seq (aka. encoder decoder) architecture can be used for sentence-level quality phrases generation. Given the sentence-level silver labels derived from our Annotator, we exploit BART, a strong generative framework with state-of-the-art performance on a variety of natural language generation (NLG) tasks, to build Generator to generate quality phrases directly.

\subsubsection{Training procedure}
Given the sentence $\textbf{s}=\{x_1, x_2, ..., x_t\}$ which contains $t$ words, and $\textbf{t}=\{k_1, k_2, ..., k_m\}$ which consists of $m$ quality phrases (silver labels). We formalize the typical phrase tagging task as a Seq2Seq generation problem by directly fine-tuning on pre-trained model BART using its conditional generation method. To this end, we define the representations for both source ($X$) and target ($Y$) sequences by applying the BART tokenizer on them. Note that $Y$ is transformed into a natural sentence by concatenating all quality phrases with a comma as a special token. These representation allow us to directly fine-tune on BART by generating all the quality phrases jointly in one decoding computation. The special token (comma) allows us to easily parse the target sequence $Y$ during the inference phrase. We fine-tune BART by maximizing the probability $P(Y|X)$ as shown in Eq.~\ref{loss_func} following the training procedures provided in the Huggingface library\footnote{https://huggingface.co/}.
\begin{equation}
    \label{loss_func}
    P(Y|X) = \prod \limits_{t=1}^{m}P(y_t|X,Y_{<t})
\end{equation}

\subsubsection{Inference procedure}
In the reference stage, our model is asked to generate up to $m$ quality phrases given an input test sentence. We adopt beam search decoding method to generate diverse quality phrases for each input sentence. To avoid uninformative phrases, we also employ the same filtering algorithm (remove stopwords and keep non-noun phrases) as we did for silver label generation in Section~\ref{segmentation}. The experimental results in Section~\ref{ablation} show that our Generator achieves higher precision scores than the strong competitors. Compared with neural discriminative models, our generative Generator requires less annotated data and is more robust to the silver labels derived from language model based Annotator (More details please see Section~\ref{robust}).

\subsection{Results Merge}
It is worth mentioning that even human annotator cannot reach complete agreement on some particular phrases. Thus, it is not surprising to see that two overlapping phrases are likely both annotated as quality phrases. In addition, as we described in the introduction, our Annotator and Generator are somewhat complementary due to their distinct characteristics. Concretely, the Annotator achieves high accuracy but relatively low recall, while the Generator does the opposite (More details please refer to Section~\ref{ablation}). These observations motivate us to merge the quality phrases derived from both Annotator and Generator as the final predictions to find more accurate and diverse quality phrases.

\section{Experimental Setup} \label{exp_setup}

To validate the effectiveness of our proposed LMPhrase, we conduct extensive experiments to answer the research questions below.

\noindent \textbf{RQ1}: Can our LMPhrase outperform the strongest baseline on sentence-level phrase tagging task?

\noindent \textbf{RQ2}: How does our LMPhrase perform for unsupervised document-level keyphrase extraction compared to strong competitors?

\noindent \textbf{RQ3}: Can our language model based Annotator generate high quality silver labels than other approaches?

\noindent \textbf{RQ4}: Whether the components of our LMPhrase (i.e. Annotator and Generator) designed for phrase mining task are effective or not?

\noindent \textbf{RQ5}: How is our LMPhrase’s ability when the amount of training data (silver labels) change?

Aiming at answering the above questions, we choose two public available benchmark datasets and adopt seven state-of-the-art baselines for a thorough comparative study. Specifically, we compare our LMPhrase with previous studies on two tasks at different granularity: sentence-level phrase tagging, and document-level keyphrase extraction.

\subsection{Datasets}

To fully evaluate all different methods, two common benchmark datasets from different domains are used. Table~\ref{tab:statistics} shows the detailed data statistics.

\noindent\textbf{KP20k} is a document collection of titles, abstracts and keyphrases from Computer Science papers~\citep{DBLP:conf/acl/MengZHHBC17}. It consists of $527,090$ for training and $20,000$ for testing.

\noindent\textbf{KPTimes} is a large dataset in which news articles are paired with keyphrases curated by editors~\citep{DBLP:conf/inlg/GallinaBD19}. In total, $259,923$ of the articles are used for training and $20,000$ of the articles were used for testing.

\begin{table}[h]
\caption{Data statistics.}
\label{tab:statistics}
\begin{tabular}{lcc}
\hline
Statistics & KP20k & KPTimes \\ \hline
\multicolumn{3}{c}{\quad \quad \quad \quad \quad \quad \quad \quad \quad \quad \quad Train Set}   \\ 
\# documents & 527,090 & 259,923 \\ 
\# words per document & 176 & 907 \\
\multicolumn{3}{c}{\quad \quad \quad \quad \quad \quad \quad \quad \quad \quad \quad Test Set}    \\
\# documents & 20,000 & 20,000 \\ 
\# multi-word keyphrases & 37,289 & 24,920 \\ 
\# unique & 24,626 & 8,970 \\ \hline
\end{tabular}
\end{table}

\subsection{Evaluation Tasks and Metrics}
\noindent\textbf{Task I: Sentence-level Phrase Tagging} is a fine-grained phrase mining task that aims to recognize quality phrases for each input sentence. We adopt the manually annotated sentence-level phrase tagging benchmark dataset released by UCPhrase~\citep{DBLP:conf/kdd/GuWBML0S21} to validate our LMPhrase framework.

We evaluate the recognized quality phrases and report the overall precision, recall and $F1$ scores. We calculate these scores in a micro-averaged manner following the method previously used on the entity recognition task~\citep{DBLP:conf/emnlp/ShangLGRR018}.

\noindent\textbf{Task II: Document-level Keyphrase Extraction} is a classic task of extracting a set of salient words or phrases from the document that can best summarize the main contents of the given document~\citep{DBLP:conf/acl/HasanN14}. Keyphrase extraction generally consists of two stage including (1) candidate keyphrase generation and (2) keyphrase importance ranking.

In the first stage, we implement the candidate keyphrase generation process by directly applying our LMPhrase on the sentences from the given document and then aggregate all the candidate quality phrases. For the ranking stage, we employ the classical Term Frequency - Inverse Document Frequency (TF-IDF) algorithm to rank the extracted candidate quality phrases. Following the standard evaluation method~\citep{DBLP:conf/jcdl/GallinaBD20}, we calculate the $F1$ score of the top-10 ranked phrases ($F1@10$) by averaging across all documents in the same dataset in a macro way to evaluate the extracted quality phrases. The document-level keyphrase extraction task is evaluated on the test sets of KP20k and KPTimes.

\subsection{Compared Methods}
We compare our LMPhrase with three types of approaches to fully validate the effectiveness of our framework in the same case, i.e., the gold labels used for training are not available. Firstly, we compare our approach with the pre-trained off-the-shelf toolkits. Then, we compare our approach with knowledge bases (KB) based distantly-supervised methods. Finally, we employ two state-of-the-art unsupervised phrase mining methods to compare with our approach.

For off-the-shelf toolkit baselines, we adopt the pre-trained linguistic-based methods.

\begin{itemize}
\item \noindent\textbf{PKE} is a commonly used keyphrase extraction toolkit. Its core module is the chunking model which is based on a supervised POS tagging model from NLTK\footnote{https://www.nltk.org/}.
\item \noindent\textbf{Spacy}\footnote{https://spacy.io/} is an industrial-grade library which is also based on the pre-trained phrase chunking model with supervised POS tagging and parsing data.
\item \noindent\textbf{StanfordCoreNLP}\footnote{https://stanfordnlp.github.io/CoreNLP/} is a widely used NLP library whose core chunking model is pre-trained on dependency parsing data.
\end{itemize}

For KB based distantly-supervised methods, the silver labels are induced from the Wiki Entities~\citep{DBLP:journals/tkde/ShangLJRVH18}.

\begin{itemize}
\item \noindent\textbf{AutoPhrase}~\citep{DBLP:journals/tkde/ShangLJRVH18} uses distant supervision for silver labels generation and leverages statistic-based binary phrase classifier with a POS-guided phrasal segmentation enhanced model to recognize quality phrase mining.
\item \noindent\textbf{Wiki+RoBERTa} is the variant of UCPhrase~\citep{DBLP:conf/kdd/GuWBML0S21} with the same transformer-based span prediction framework, but using distant supervision for silver labels generation instead.
\end{itemize}

For unsupervised methods we consider:
\begin{itemize}
\item \noindent\textbf{ToPMine}~\citep{DBLP:journals/pvldb/El-KishkySWVH14} is a strong unsupervised phrase mining approach building upon statistical features. It presents a topical phrase mining framework to discover arbitrary length topical phrases.
\item \noindent\textbf{UCPhrase}~\citep{DBLP:conf/kdd/GuWBML0S21} is the state-of-the-art unsupervised phrase tagging approach. It induces quality phrase as silver labels by extracting maximal word sequences from the given document, then trains a transformer-based span classification model to predict quality phrases.
\end{itemize}

\begin{table*}[t!]
\caption{Evaluation results (\%) of sentence-level phrase tagging for all compared methods. The best results are in bold, and the second best are underlined.}
\label{tab:sentence_level}
\begin{tabular}{cccccccc}
\hline
\multirow{2}{*}{Model Type} & \multirow{2}{*}{Model Name} & \multicolumn{3}{c}{KP20k} & \multicolumn{3}{c}{KPTimes} \\ 
 &  & \multicolumn{1}{c}{Precision} & \multicolumn{1}{c}{Recall} & F1 & \multicolumn{1}{c}{Precision} & \multicolumn{1}{c}{Recall} & F1 \\ \hline
\multirow{3}{*}{Off-the-shelf Toolkit} & PKE & \multicolumn{1}{c}{54.1} & \multicolumn{1}{c}{63.9} & 58.6 & \multicolumn{1}{c}{56.1} & \multicolumn{1}{c}{62.2} & 59.0 \\ 
 & Spacy & \multicolumn{1}{c}{56.3} & \multicolumn{1}{c}{68.7} & 61.9 & \multicolumn{1}{c}{61.9} & \multicolumn{1}{c}{62.9} & 62.4 \\ 
 & StanfordNLP & \multicolumn{1}{c}{48.3} & \multicolumn{1}{c}{60.7} & 53.8 & \multicolumn{1}{c}{56.9} & \multicolumn{1}{c}{60.3} & 58.6 \\ \hline
\multirow{2}{*}{Distantly Supervised} & AutoPhrase & \multicolumn{1}{c}{55.2} & \multicolumn{1}{c}{45.2} & 49.7 & \multicolumn{1}{c}{44.2} & \multicolumn{1}{c}{47.7} & 45.9 \\ 
 & Wiki+RoBERTa & \multicolumn{1}{c}{58.1} & \multicolumn{1}{c}{64.2} & 61.0 & \multicolumn{1}{c}{60.9} & \multicolumn{1}{c}{65.6} & 63.2 \\ \hline
\multirow{3}{*}{Unsupervised} & TopMine & \multicolumn{1}{c}{39.8} & \multicolumn{1}{c}{41.4} & 40.6 & \multicolumn{1}{c}{32.0} & \multicolumn{1}{c}{36.3} & 34.0 \\ 
 & UCPhrase & \multicolumn{1}{c}{\underline{69.9}} & \multicolumn{1}{c}{\underline{78.3}} & \underline{73.9} & \multicolumn{1}{c}{\underline{69.1}} & \multicolumn{1}{c}{\underline{78.9}} & \underline{73.5} \\  
 & LMPhrase (Ours) & \multicolumn{1}{c}{\textbf{71.9}} & \multicolumn{1}{c}{\textbf{79.2}} & \textbf{75.3} & \multicolumn{1}{c}{\textbf{73.2}} & \multicolumn{1}{c}{\textbf{79.2}} & \textbf{76.1} \\ \hline
\end{tabular}
\end{table*}

\subsection{Implementation Details}
For silver label acquisition, \textit{Stanford CoreNLP Tool} is employed for sentence tokenizing and part-of-speech tagging. We employ regular expression $\{\langle NN.\ast | JJ \rangle \ast \langle NN. \ast \rangle\}$ to recognize noun phrases, where $NN$ and $JJ$ denote the POS tags of noun and adjective, $*$ acts as the wildcard character. We use the uncased RoBERTa-base version\footnote{https://huggingface.co/roberta-base} and SciBERT\footnote{https://huggingface.co/allenai/scibert\_scivocab\_uncased} as the base pre-trained language model of Annotator for KPTimes and KP20k, respectively. They will be used for silver label acquisition.

We randomly selected a portion of the training set and manually annotated it to create a validation set. Subsequently, we conducted a pilot study on how to determine the optimal hyper-parameter threshold value of the segmentation algorithm. We found empirically that setting the threshold of the segmentation algorithm to $40th$ percentile of all the adjacent words correlation values sequence of the given input sentence gave the best performance on the validation set. 

For quality phrase generation, we randomly split the silver labels mined from Annotator into training and validation sets with a ratio of $8$:$2$. The number of training set is a hyper-parameter that affects the performance of the Generator. We analyze the influence of this hyper-parameter and report the optimal settings for each dataset. For more details on the sensitivity analysis of this parameter please refer to Section~\ref{robust}. We fine-tune the conditional generation task on BART-large model\footnote{https://huggingface.co/facebook/bart-large} for a maximum of $50$ epochs on a GeForce GTX-1080Ti GPU, retaining the checkpoint with the highest $F1$ score on the validation set. We adopt the default dynamic learning rate, warm up $1000$ iterations, and decay afterward. The batch size is to $8$.

\section{Experimental Results} \label{exp_result}

\subsection{Results of sentence-level phrase tagging (RQ1)} 
Table~\ref{tab:sentence_level} shows the sentence-level phrase tagging performance of our approach and the compared baselines on two benchmark datasets. As can be seen, our LMPhrase consistently outperforms all the compared methods in terms of overall performance, achieving new state-of-the-art performance on all two datasets. These promising results validate the effectiveness our framework for the sentence-level phrase tagging task. In detail, our LMPhrase outperforms the strongest baseline UCPhrase by $1.4\%$ and $2.6\%$ in $F1$ on KP20k and KPTimes datasets, respectively. The large margin gains by LMPhrase through combining both Annotator and Generator strongly demonstrate that our framework is very effective for this task. 

Concretely, the encouraging results achieved by LMPhrase are mainly attributed to the following two aspects: (1) Our proposed silver label Annotator benefits from the rich knowledge from pre-trained language models in a context-aware manner, and thus offer the advantages in preserving informativeness,
concordance, and completeness of quality phrases. (2) Fine-tuning on strong Seq2Seq generative pre-trained model BART with quality silver labels mined by Annotator is very robust and is able to generate more accurate and complete quality phrases.

\subsection{Results of document-level keyphrase extraction (RQ2)} 
Compared to fine-grained sentence-level phrase tagging, keyphrase extraction is a relatively coarse phrase mining task (i.e., document-level). It first requires model to have a comprehensive understanding of the document and then extract several salient phrases that can best summarize the main contents of the given document, all of which pose a great challenge to the model.

We directly apply our LMPhrase framework to the candidate keyphrase generation stage of the keyphrase extraction task. The comparative results of all the methods for the unsupervised document-level keyphrase extraction task on two benchmark datasets are presented in Table~\ref{tab:document_level}. From the results, we can find that our LMPhrase outperforms all the compared methods on $F1@10$ score. The encouraging results indicate the effectiveness of LMPhrase and show its great value to the application of unsupervised keyphrase extraction.

\begin{table}[t!]
\caption{Evaluation results (F1@10 \%) of document-level keyphrase extraction for all compared methods. The best results are in bold, and the second best are underlined.}
\label{tab:document_level}
\centering
\resizebox{0.95\linewidth}{!}{
\begin{tabular}{cccc}
\hline
Model Type & Model Name & KP20k & KPTimes \\ \hline
\multirow{3}{*}{Off-the-shelf Toolkit} & PKE & 12.6 & 4.4 \\ 
 & \multicolumn{1}{c}{Spacy} & 15.3 & 8.6 \\ 
 & \multicolumn{1}{c}{StanfordNLP} & 13.9 & 8.7 \\ \hline
\multirow{2}{*}{Distantly Supervised} & AutoPhrase & 18.2 & 10.3 \\  
 & Wiki+RoBERTa & 19.2 & 9.4 \\ \hline
\multirow{3}{*}{Unsupervised} & TopMine & 15.0 & 8.5 \\  
 & \multicolumn{1}{c}{UCPhrase} & \underline{19.7} & \underline{10.9} \\  
 & \multicolumn{1}{c}{LMPhrase (Ours)} & \textbf{21.5} & \textbf{11.4} \\ \hline
\end{tabular}}
\end{table}

\subsection{Evaluation of silver labels acquisition (RQ3)} ~\label{Silver_Labels}
We also explore our Annotator’s ability in mining quality phrases as silver labels from the given sentence. To this end, we adopt two alternative approaches for silver label acquisition, namely, NPChunk (noun phrase chunking) and MaxSpan (maximum span from all possible spans), which are commonly used in the generation of candidate keyphrases generation phase of unsupervised keyphrase extraction model. Specifically, NPChunk employs StanfordCoreNLP Tools and regular expression to identify noun phrases\footnote{we extract noun phrases based on the regular expression which encodes the phrase pattern of interest $\{ \langle NN.\ast | JJ\rangle \ast \langle NN. \ast \rangle\}$, where $NN$ and $JJ$ denote the  POS tags of noun and adjective, $*$ acts as the wildcard character.} as candidate quality phrases according to our preliminary analysis in Section~\ref{quality_analysis}. MaxSpan keeps the maximum span sequence as candidate quality phrase if there exist overlaps among all possible multi-word spans. We also apply the same simple filtering algorithm (see Section~\ref{segmentation}) before finalizing the silver labels to avoid uninformative candidate phrases as our Annotator does.

We compare the results of silver labels acquisition mined by different approaches on sentence-level phrase tagging task. As illustrated in Table~\ref{tab:annotator}, it can be seen that our Annotator outperforms both NPChunk and MaxSpan in terms of $F1$ on all two datasets, which verifies the effectiveness of our Annotator. Indeed, while NPChunk and MaxSpan generally achieves better recall scores compared with our Annotator, their precision scores are much lower, which hinders the overall performance. The reason might be that a significant portion of noun phrases are not quality phrases. (e.g. 'rapid development', 'task types', 'significant applications', etc.) Our Annotator benefits from huge syntactic, semantic and structural from large pre-trained language model and is therefore able to mine higher quality phrases as silver labels.

\begin{table}[]
\caption{Evaluation results of silver labels on two benchmark datasets for sentence-level phrase tagging task. The best results are in bold.}
\label{tab:annotator}
\centering
\resizebox{0.95\linewidth}{!}{
\begin{tabular}{ccccccc}
\hline
\multirow{2}{*}{Models} & \multicolumn{3}{c}{KP20k} & \multicolumn{3}{c}{KPTimes}  \\ \cline{2-7}
 & Precision & Recall & F1 & Precision & Recall & F1  \\ \hline
NPChunk  &  58.8 & 72.5 & 64.9 & 60.5 & \textbf{75.1} & 67.0 \\  
MaxSpan  &  30.0 & \textbf{72.6} & 42.5 & 24.9 & 74.3 & 37.3 \\ 
Annotator  &  \textbf{72.4} & 64.4 & \textbf{68.1} & \textbf{74.7} & 67.9 & \textbf{71.1} \\ \hline
\end{tabular}}
\end{table}

\subsection{Ablation studies (RQ4)} ~\label{ablation}
To assess the rationality of our LMPhrase, we conduct ablation study to evaluate the contribution of Annotator and Generator, respectively. From the Table~\ref{tab:ablation}, we can find the performances drop if only Annotator or Generator is used separately. Note that our Generator achieves the best precision values when compared to other models, which validates the strong generative power and robustness of our Generator. We envision that our LMPhrase can be further improved if more data are given.

Given that even human annotators may not fully agree with each other on some specific phrases, it leads to more diverse quality phrases which are very likely annotated as gold quality phrases. In addition, considering the distinct characteristic of Annotator and Generator, LMPhrase (i.e., combination of Annotator and Generator) is able to extract more diverse and complete quality phrases, resulting more significant recall. Thus it is not surprising to see LMPhrase performs the best. The experimental observation further verifies the complementary nature of Annotator and Generator. In Section~\ref{case_study}, we provide a comprehensive comparison of the different components of LMPhrase (i.e., Annotator and Generator) on real-word examples to get a more straightforward visualization of the advantages and disadvantages of Annotator and Generator.

\begin{table}[]
\caption{Evaluation results of ablation experiments on two benchmark datasets for sentence-level phrase tagging task. The best results are in bold.}
\label{tab:ablation}
\centering
\resizebox{0.95\linewidth}{!}{
\begin{tabular}{ccccccc}
\hline
\multirow{2}{*}{Models} & \multicolumn{3}{c}{KP20k} & \multicolumn{3}{c}{KPTimes}  \\ \cline{2-7}
& Precision & Recall & F1 & Precision & Recall & F1  \\ \hline
Annotator  &  72.4 & 64.4 & 68.1 & 74.7 & 67.9 & 71.1 \\ 
Generator  &  \textbf{78.2} & 63.5 & 70.1 & \textbf{79.6} & 61.7 & 69.5 \\  
LMPhrase (Ours)  &  71.9 & \textbf{79.2} & \textbf{75.3} & 73.2 & \textbf{79.2} & \textbf{76.1} \\ \hline
\end{tabular}}
\end{table}

\subsection{Impact of the size of Silver Labels (RQ5)} ~\label{robust}
\begin{figure*}[]
    \centering
    \includegraphics[scale=0.55]{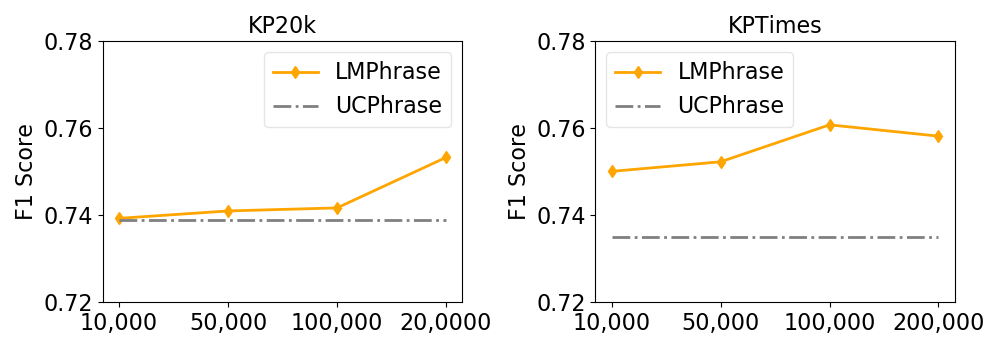}
    \caption{Evaluation results of sentence-level phrase tagging on two benchmark datasets with varying number of silver labels.}
    \label{fig:impact_f1}
\end{figure*}

We access the performance of our LMPhrase on two datasets when the number of training data (i.e., silver labels) varies. As demonstrated in Figure~\ref{fig:impact_f1}, our LMPhrase achieves continuous performance improvements on the two datasets as the size of training data increases. Note that with only $10,000$ training examples (i.e., sentences with silver labels mined from our Annotator) for both KP20k and KPTimes, our LMPhrase achieves results on par (even better for KPTimes) with the strongest UCPhrase trained on full data. This result demonstrates the ability of LMPhrase in generating quality phrases. On one hand, it indicates the higher quality of silver labels mined by our Annotator. On the other hand, it validates the effectiveness and robustness of our powerful pre-trained language model based Generator. We speculate that the possible reason for slight decrease for $200,000$ on KPTimes is that $100,000$ training sample might be enough for KPTimes, adding more sentences is less helpful.

\subsection{Qualitative Analysis} \label{case_study}
\begin{figure*}[]
    \centering
    \includegraphics[scale=0.45]{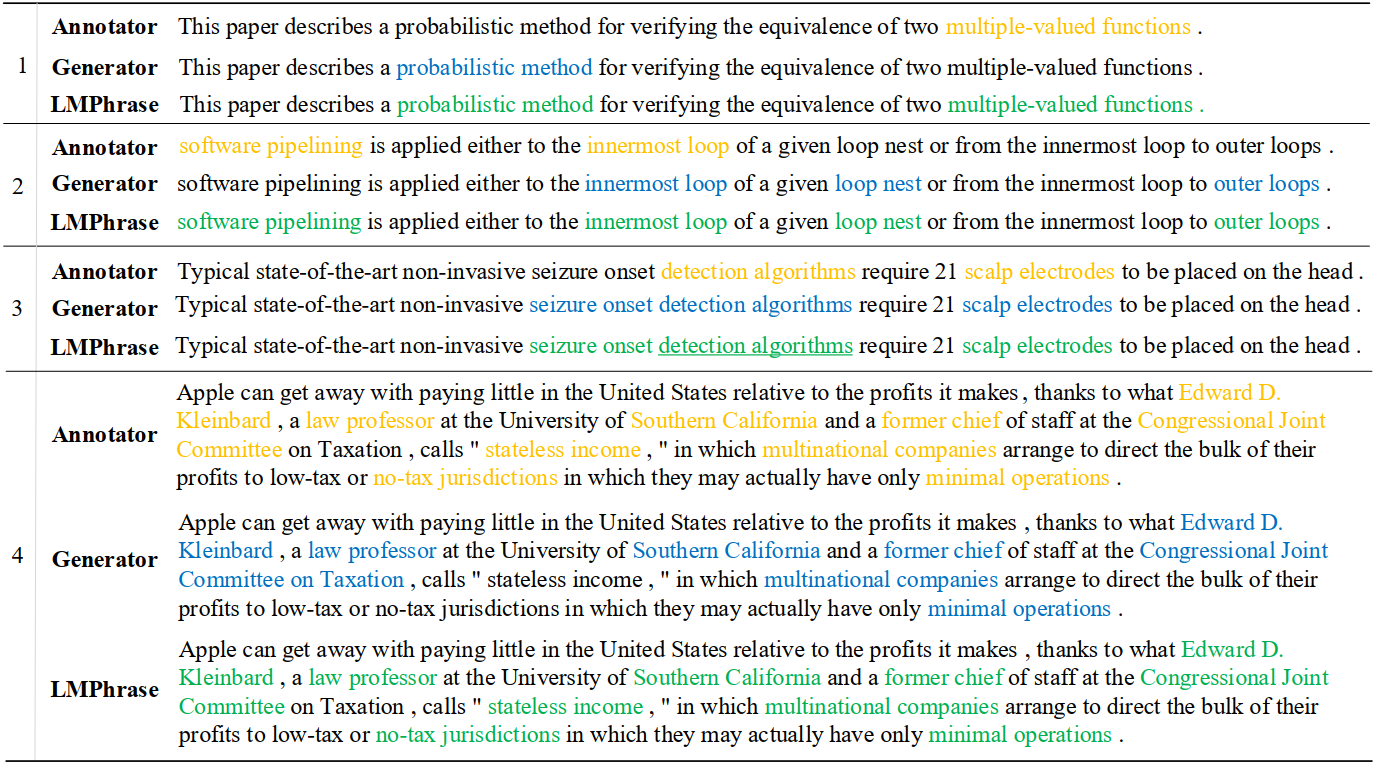}
    \caption{Qualitative Analysis.}
    \label{fig:case}
\end{figure*}

We conduct a case study to further validate the effectiveness of our framework. Concretely, we select four representative sentences from the test set and show the predictive results of our approach and the ablates in Figure~\ref{fig:case}. As we can see, for the first two examples, both Annotator and Generator can only extract partial quality phrases from the given sentence. It also indicates that the quality phrases derived from these two components are complementary to some extent. For case 3 where the gold quality phrases are overlapped, in other words, human annotators may not fully agree with each other on certain phrases. For instance, 'seizure onset detection algorithms' and 'detection algorithm' are both annotated as gold quality phrases. However, these two quality phrases are successfully extracted by our Generator and Annotator, respectively, which verify the reasonableness of the result merge. As to the last case which contains nine quality phrases including named entity, terminology, emerging phrase and acronyms, which poses a significant challenge to the model. Nevertheless, our LMPhrase is still able to extract all the quality phrases while separate Annotator and Generator failed. These examples further validate that our context-aware LMPhrase is well suited for the phrase mining task.

\section{Conclusion} \label{conclusion}
In this paper, we propose LMPhrase, a novel unsupervised context-aware quality phrase mining framework with large pre-trained language model. Specifically, we design Annotator to mine quality phrases as silver labels by employing Perturbed Masking technique on the auto-encoding pre-trained language model BERT. Then the encoder-decoder pre-trained language model BART is adopted as an Generator to recognize the quality phrases directly. Lastly, we merge the results of quality phrases derived from both Annotator and Generator as the final predicted quality phrases.

We have conducted extensive experiments to validate and reveal the strength of our two major components: our Annotator considers the contextual information, thus offer the advantages in preserving informativeness, concordance, and completeness of quality phrases. Our Generator, which is fine-tuned on pre-trained language model BART, remains effective and robust even with limited annotated resources. Future work will consider the applications of our LMPhrase in other natural language processing tasks. We will also try to further improve the model performance of Annotator by incorporating task-specific self-supervised learning.

\bibliographystyle{cas-model2-names}

\bibliography{cas-refs}

\end{document}